\documentclass[runningheads]{llncs}
\usepackage{graphicx}
\graphicspath{{fig/}} 
\usepackage{amsmath}
\usepackage{hyperref} 
\usepackage{bm}

\usepackage{subcaption}
\usepackage{xcolor}

\begin{document}
\title{The Inefficiency of Genetic Programming for Symbolic Regression\thanks{This is an extended version of the article Gabriel Kronberger, Fabricio Olivetti de Franca, Harry Desmond, Deaglan J. Bartlett, and Lukas Kammerer. 2024. The Inefficiency of Genetic Programming for Symbolic Regression. In Parallel Problem Solving from Nature – PPSN XVIII: 18th International Conference, PPSN 2024, Hagenberg, Austria, September 14--18, 2024, Proceedings, Part I. Springer-Verlag, Berlin, Heidelberg, 273--289. https://doi.org/10.1007/978-3-031-70055-2\_17}}
\titlerunning{The Inefficiency of GP for SR}
\author{Gabriel Kronberger\inst{1}\orcidID{0000-0002-3012-3189} \and
Fabricio Olivetti de Franca\inst{2}\orcidID{0000-0002-2741-8736} \and
Harry Desmond\inst{3}\orcidID{0000-0003-0685-9791}\and 
Deaglan J. Bartlett\inst{4}\orcidID{0000-0001-9426-7723} \and 
Lukas Kammerer\inst{1}\orcidID{0000-0001-8236-4294}}
\authorrunning{G. Kronberger et al.}
\institute{University of Applied Sciences Upper Austria, Softwarepark 11, 4232 Hagenberg, Austria
\email{gabriel.kronberger@fh-ooe.at}\\
\and
Federal University of ABC, Santo Andre, SP, Brazil
\email{folivetti@ufabc.edu.br}\\
\and
Institute of Cosmology \& Gravitation, University of Portsmouth, Dennis Sciama Building, Portsmouth, PO1 3FX, UK\\
\and
CNRS \& Sorbonne Universit\'e, Institut d’Astrophysique de Paris (IAP), UMR 7095, 98 bis bd Arago, F-75014 Paris, France
}
\maketitle              
\begin{abstract}
We analyse the search behaviour of genetic programming for symbolic regression in practically relevant but limited settings, allowing exhaustive enumeration of all solutions. This enables us to quantify the success probability of finding the best possible expressions, and to compare the search efficiency of genetic programming to random search in the space of semantically unique expressions. 
This analysis is made possible by improved algorithms for equality saturation, which we use to improve the Exhaustive Symbolic Regression algorithm; this produces the set of semantically unique expression structures, orders of magnitude smaller than the full symbolic regression search space. We compare the efficiency of random search in the set of unique expressions and genetic programming.
For our experiments we use two real-world datasets where symbolic regression has been used to produce well-fitting univariate expressions: the Nikuradse dataset of flow in rough pipes and the Radial Acceleration Relation of galaxy dynamics.
The results show that genetic programming in such limited settings explores only a small fraction of all unique expressions, and evaluates expressions repeatedly that are congruent to already visited expressions.  

\end{abstract}

\section{Introduction}

Symbolic regression (SR) is a machine learning task in which the goal is to find an expression from a parametric function family $f(x; \theta)$ that accurately fits a dataset $(x_i, y_i)_{i=1}^n$ by adjusting $\theta$~\cite{koza1992,kronberger2024}.
This is commonly performed using search algorithms that explore a search space composed of mathematical expressions. 

Search algorithms~\cite{norvig2002modern} can be classified as complete and optimal if they are guaranteed to find a solution and also return the best possible solution.
A random search, in which a random solution is sampled at every time step, is both complete and optimal, provided an unlimited runtime.
Another complete and optimal search is the enumeration of the whole search space, since the algorithm will traverse all of the possible solutions. This is only feasible if the search space is finite and small enough to process all elements in limited time.
On the other hand, a local search exploits the neighbourhood of an initial point and focuses the search on a limited region of the search space. This approach is neither complete nor optimal as it depends on the size of the explored neighbourhood and the starting point.

In evolutionary algorithms (EA)~\cite{eiben2015introduction}, a set of solution candidates (here called a population) is used in a parallel search. Parts of solution candidates are repeatedly recombined, mutated, and evaluated to gradually evolve improved solutions. This process simulates mechanisms observed in natural evolution, most importantly inheritance of traits and selection pressure.
Genetic programming (GP)~\cite{koza1992} is an evolutionary algorithm and a popular method for SR. In tree-based GP for SR, the \emph{search space} of genotypes (i.e., expression trees that can be generated up to a given length limit) can be distinguished from the much smaller set of distinct expressions after conversion to canonical form, because many different expressions represent the same function (e.g., $x\,(x+p_1)$ and $p_1\,x + x^2$). 
Many modern GP systems use a solver for more efficient parameter optimisation~\cite{kommenda2020parameter,burlacu2020operon}. Expressions may therefore contain placeholders ${p_1} \ldots p_k$ for real-valued fitting parameters, and reparameterised forms can represent the same function (e.g., $p_1\,(x + p_2)$ and $p_1\, x + p_2'$). 
The \emph{solution space} is the space of functions that can be produced by fitting the elements from the search space to a given dataset. The elements of the solution space include the best-fit parameter values and the fitting objective value (e.g. log-likelihood for maximum likelihood, or the mean of squared errors for least-squares fitting). 
As a consequence of local optima when fitting parameters, GP systems may produce different solutions for the same expression. 

The efficiency of a search algorithm can be quantified via the probability of reaching a solution of a certain quality by the number of evaluated solution candidates.
An efficient search algorithm samples solution candidates with minimal repetition, and invests more time evaluating more promising solution candidates. Ideally, it enumerates the solution candidates following an order from the most promising to the least promising, thus hopefully reaching an acceptable solution quickly.
For SR, this can be challenging because there are many equivalent representations of the same function, either because they have the same fitness or because they are isomorphic modulo parameter values. Whether this helps to traverse SR search spaces is still debated. Some authors argue that it degrades overall performance, while others show that it may be important for reaching a global optimum, since neutral steps can be necessary to reach promising regions\cite{ebner1999search,hu2018neutrality,hu2023phenotype,banzhaf2024combinatorics}.

In this paper, we analyse the efficiency of GP for SR with parameter optimisation~\cite{kommenda2020parameter}. We define efficiency as the success probability after a given number of evaluated solution candidates. We propose a method that explicitly quantifies how many evaluated candidates are unique expressions, or alternatively, how often GP revisits expressions that are isomorphic to previously evaluated ones. In contrast to previous work, we do not use fitness values to detect semantic duplicates, because fitness depends on optimised parameter values and can therefore only detect semantically identical expressions when they are parameterised equivalently. Instead, we use symbolic simplification to a canonical form, which lets us detect duplicate expression forms regardless of parameterisation.
We use equality saturation~\cite{de2023reducing,willsey2021egg}, adapted to simplify and rewrite equivalent expressions into the same form, so we can measure how many isomorphic solutions are visited throughout the search. The analysis is done for a limited search space, for which we fully enumerate all solutions using Exhaustive Symbolic Regression (ESR)~\cite{bartlett2023exhaustive} for two practically relevant datasets from the physical sciences. In this way, we can calculate the success probability of finding the best solution or, for example, one of the top 100 solutions with GP. 

The results show a worrying inefficiency of GP for SR with a very low rate of unique solutions and a success probability smaller than an idealised random search when limiting the search space to short expressions.

In the following sections, we review related work (Sec.~\ref{sec:related}), describe our methods (Sec.~\ref{sec:methods}), and present the results (Sec.~\ref{sec:results}). We then discuss limitations (Sec.~\ref{sec:limits}) and conclude (Sec.~\ref{sec:conclusion}).

\section{Related work}
\label{sec:related}
The characteristics of the GP search space and the biases of GP when exploring the search space have been extensively studied in earlier work.

In early work, Ebner~\cite{ebner1999search} investigated the redundancy in the solution encoding in GP for SR and argued that, even though the search space of SR is much larger than commonly seen in genetic algorithms, this redundancy is important to guarantee that one of such equivalent solutions is reached.
Later, Daida et al.~\cite{daida2003makes,daida2003identifying} asked the question ``What makes a GP problem hard?'' and studied the consequences of using an iterative growth mechanism for trees in GP. They showed that certain tree structures are much more likely to be visited during the exploration of the search space than others. If a solution lies within the unlikely region of the search, GP has a lower success rate.

Gustafson et al.~\cite{gustafson2005improving} observed a high frequency of offspring sharing the same fitness values as one of their parents, even with structural differences, chiefly when both parents had a similar fitness value. They suggested to disallow the recombination of two parents with the same fitness value. With this simple mechanism, they observed a significant increase in improved and worsened offspring, and a high decrease of no-change offspring.

Neutrality and redundancy have further been studied by Hu, Banzhaf, and Ochoa \cite{hu2018neutrality,hu2023phenotype,banzhaf2024combinatorics} for linear GP for Boolean SR problems with the help of search trajectory networks. 
They found that more complex phenotypes are harder for evolution to discover as they are represented by fewer genotypes and overrepresented  phenotypes are easier to find. 

In~\cite{niehaus2007reducing}, it was shown that caching fitness values and simplification of solution candidates can significantly reduce the evaluation time for graph-based GP. This observation implies that GP generates a significant number of expressions that can be simplified to the same expression. Similarly, McPhee et al.~\cite{mcphee2008semantic} analysed the possible semantic outcomes of subtree crossover and found that for Boolean GP, most crossover events (over 75~\%) produced no immediately useful semantic changes, drawing the attention to the need of investigating new crossover operators for GP.

More recently, Langdon~\cite{Langdon2021} showed that function-evaluation values and opcodes are similar or identical in expanding regions from the root node in SR trees, including large trees over thousands of GP generations. Nevertheless, fitness continued to evolve even though most crossover events had near-zero disruption.

Simplification of SR expressions during the evolution or creation of the expressions have been shown to be beneficial ~\cite{rivero2022dome,randall2022bingo,cao2023genetic,seidyo2024inexact} as well as an important mechanism for diversity control~\cite{burlacu2019online,burlacu2020hash}.

Several alternative algorithms for SR have been proposed, which try to improve efficiency by preventing re-evaluations of redundant expressions. This includes algorithms that enumerate a restricted search space, such as~\cite{worm2013prioritized,kammerer2020symbolic,bartlett2023exhaustive} or algorithms that decompose the problem and build the solutions incrementally \cite{de2018greedy,rivero2022dome}.
Others use an evolutionary approach but restrict the search space to find less complex models~\cite{virgolin2017scalable,de2021interactionB,de2022transformation,kartelj2023rils}.

Exhaustive Symbolic Regression~\cite{bartlett2023exhaustive} (ESR) is an SR algorithm that affords enumeration of the full SR solution space as defined by the function and terminal sets and a length limit. The algorithm as described by \cite{bartlett2023exhaustive} has three phases. First it generates all valid expression trees. In the second phase the expression trees are simplified using SymPy, a Python software library for symbolic computation, and only unique expressions are kept. Finally, the parameters of the unique expressions are optimized to fit the expressions to a given dataset. The first two phases only need to be executed once for a given terminal and function set as they are independent of the dataset. Only the last phase of parameter optimization has to be done for each dataset individually. 

Compared to GP, ESR guarantees to find the best solution -- assuming the optimal parameters are found. It fits and evaluates only simplified expressions to prevent evaluations of structurally different but semantically identical expressions. 
However, the runtime grows exponentially with the number of variables and the maximum length limit. 

\section{Methods}
\label{sec:methods}

\subsection{Improved Exhaustive Symbolic Regression}

To allow enumeration of the search space, we limit the set of operators to $\{+, -, \cdot, \div, x^{-1}, \text{powabs}(x,y)\}$
and the expression length, defined as the number of operator and operand nodes in the expression tree, to at most $\{10, 12, 20\}$. 
For the length limits of 10 and 12 nodes, we exhaustively generate the solution space with ESR, the limit of 20 nodes is only used with GP.

Based on the original Python implementation of ESR\footnote{\url{https://github.com/DeaglanBartlett/ESR}}, we prepared a new implementation in Julia with several improvements.
In our improved implementation, expressions are generated by repeatedly applying derivation rules from a formal grammar using a breadth-first search procedure. The grammar $G(\text{E})$ used for this work is:
\begin{align*}
  \text{E} & \rightarrow x \hspace{0.5em} |  \hspace{0.5em} p  \hspace{0.5em} | \hspace{0.5em}  \text{inv}\, `(`\, \text{E}\,`)` \hspace{0.5em}  | \hspace{0.5em}  \text{powabs}\, `(`\, \text{E,\, E}\,`)` \hspace{0.5em} | \hspace{0.5em} `(`\, \text{E}\hspace{0.5em} (+|-)\hspace{0.5em} \text{E}\, `)` \hspace{0.5em}  | \hspace{0.5em} \text{E}\hspace{0.5em} (\times|\div)\hspace{0.5em} \text{E}
\end{align*}
whereby $\text{inv}(a) \equiv a^{-1}$, $\text{powabs}(a,b) \equiv |a|^b$, and $p$ is a placeholder for free parameters.

Breadth-first search starts with a sequence containing only the root symbol. In each iteration, one sequence is taken from the queue, and the first occurrence of the non-terminal symbol E (for expression) is replaced with all grammar alternatives that fit within the length limit. If the generated derivation contains only terminal symbols, the expression is complete. All other derivations are enqueued again for later processing, after checking for semantic duplicates.
Sequences that are semantically equivalent to an earlier visited sequence are detected and removed. This semantic de-duplication is implemented via simplification of expressions to a canonical form using equality graphs (e-graph), which is a data structure that allows compact storage of expressions that are congruent with respect to a given set of rules~\cite{nelson1980techniques,willsey2021egg}. 
E-graphs were originally developed for automated theorem proving. More recently, the equality saturation (eq-sat) algorithm for generating all equivalent expressions from a set of rules was improved~\cite{willsey2021egg} and released as an easy-to-use Rust library, leading to substantial efficiency improvements in several applications~\cite{willsey2021egg}.
In the context of SR, equality saturation (eq-sat) provides a mechanism for simplifying expressions to a canonical form, which opens many new pathways for more detailed study of SR algorithms and their improvement. For example, we have used eq-sat for simplifying solutions produced by GP to remove redundant parameters~\cite{de2023reducing,kronberger_jsc_2024_to_appear}. 

In this work, we use the Julia implementation provided by the package \emph{MetaTheory.jl} and the rule set shown in Table~\ref{tab:eq-rule-set}, which mainly includes commutativity, associativity, and distributivity rules for the operator set used in ESR. Eq-sat provides functionality for tracking semantic analysis values for equality classes, which we use to fold constant expressions and expressions without variables, i.e., expressions that contain only parameters and constants and can therefore be simplified to a single parameter.

After a limited number of eq-sat iterations, we extract from the e-graph the expression with the minimum number of parameters, then the minimum number of tree nodes, and finally the first one according to a recursively defined order relation. The extracted expression is the canonical representative of all congruent expressions under the rule set. We calculate a hash value for this expression and use it for semantic de-duplication, which allows us to prune redundant branches of the search tree. 
The efficiency of eq-sat enables us to generate all unique expressions for the grammar up to a maximum length of 12 within 5 to 10 hours of runtime (single-core) on a desktop PC.

\begin{table}
\centering
\setlength{\tabcolsep}{3pt} 
\caption{Rule set used for equality saturation. Symbols $a,b,c,d$ represent any expression.}
\label{tab:eq-rule-set} 
\begin{tabular}{rcll|lrcl}
$0 + a$      & $\rightarrow$  & $a$                        & $\quad$&$\quad$ & $\text{abs}(a) $ & $\rightarrow$ & $ a\quad |\quad a \leq 0$                               \\
$0 \times a$ & $\rightarrow$  & $0$                        &        &        & $\text{abs}(-1 \times a) $ & $\rightarrow$ & $ \text{abs}(a)$                              \\
$0 / a $ & $\rightarrow$ &   $0$                           &        &        & $\text{abs}(a - b) $ & $=$ & $ \text{abs}(b - a)$                                          \\
$1 \times a $ & $\rightarrow$ & $a$                        &        &        & & &                                                                                        \\
$a - a $ & $\rightarrow$ & $0$                             &        &        & $a + b\times a $ & $\rightarrow$ & $ (1 + b) \times a$                                     \\
$a + a $ & $\rightarrow$ & $2\, a$                         &        &        & $b \times a + c \times a $ & $=$ & $ a \times (b + c)$                                     \\
$a / a $ & $\rightarrow$ & $1$                             &        &        &                           & &                                                              \\
$1 ^ a $ & $\rightarrow$ & $1$                             &        &        & $a + b \times c $ & $\rightarrow$ & $ (a/b + c) \times b$                                  \\
$a ^ 1 $ & $\rightarrow$ & $a$                             &        &        & $a \times c + b \times d $ & $\rightarrow$ & $  b \times (a/b \times c + d )$              \\
$a ^ 0 $ & $\rightarrow$ & $ 1$                            &        &        & & &                                                                                        \\
$0 ^ a $ & $\rightarrow$ & $0\quad | \quad   a > 0$        &        &        & $a ^ b \times a ^ c $ & $\rightarrow$ & $ a ^ {b+c}$                                       \\
        &  &                                               &        &        & $(a^b)^c $ & $\rightarrow$ & $  a^{b\times c} \quad | \quad \text{is\_integer}(c)$         \\
$a + b $ & $=$ & $ b + a $                                 &        &        & $(a\times b)^c $ & $\rightarrow$ & $  a^c \times b^c \quad | \quad \text{is\_integer}(c)$  \\
$a \times b $ & $=$ & $ b \times a$                        &        &        & $a^c \times b^c $ & $\rightarrow$ & $  (a\,b)^c \quad | \quad \text{is\_integer}(c)$       \\
$a + (b + c) $ & $=$ & $ (a + b) + c$                      &        &        & $a \times a $ & $=$ & $ a ^ 2$                                                             \\
$a \times (b \times c) $ & $=$ & $ (a \times b) \times c$  &        &        & $(a ^ b) \times a $ & $\rightarrow$ & $ a ^ {1 + b}$                                       \\
      & &                                                  &        &        &       &  &                                                                                 \\
$a / b $ & $=$ & $ a\times b^{-1}$                         &        &        &       &  &                                                                                 \\
$(1 / a)\times b $ & $\rightarrow$ & $ b / a$              &        &        &       &  &                                                                                 \\
$-(-a) $ & $\rightarrow$ & $ a$                            &        &        &       &  &                                                                                 \\
$-a $ & $=$ & $ (-1)\times a$                              &        &        &       &  &                                                                                 \\
$a - b $ & $=$ & $ a + (-1)\times b$                       &        &        &       &  &                                                                                
\end{tabular}
\end{table}

We have adapted the implementation of eq-sat as described in~\cite{de2023reducing} to simplify the expressions while minimizing the number of parameters and returning a canonical representation of the expression. By this, we can map isomorphic expressions into a single hash value, helping in the detection of redundant genotypes, similar to ~\cite{burlacu2019online}. Notice that eq-sat only guarantees the detection if we have a sufficient set of rewriting rules and the e-graphs are saturated. For the purpose of this work, we stop after a fixed number of eq-sat iterations which allows us to detect most of the duplicates. 

\begin{figure}[t!]
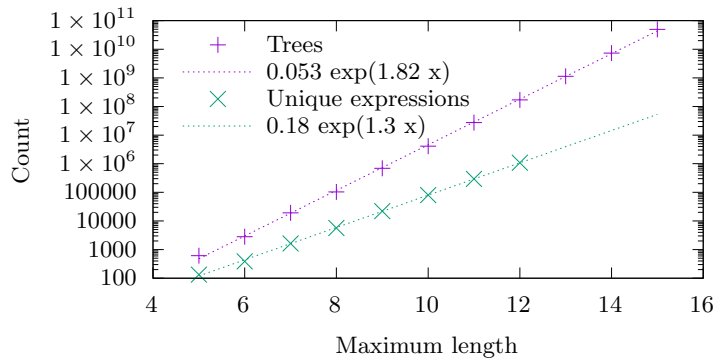

    \centering
    \include{fig/sr-scaling}
    \caption{Growth of the search space and solution space sizes for the used function set. After simplification we found $80\,407$ unique expressions with maximum length $10$, and $1\,083\,803$ unique expressions with maximum length $12$. At maximum length 10 there are approximately 50$\times$ more trees than unique expressions. This factor grows exponentially with maximum length. }
    \label{fig:search-space-size}
\end{figure}

Figure~\ref{fig:search-space-size} shows the exponential growth of the number of expression trees and unique simplified expressions with the expression length limit and the grammar $G(E)$. The gap between the two values also grows exponentially, which implies that the redundancy problem becomes more extreme for longer length limits.

The final step of ESR is fitting all unique expressions to the dataset, which means finding optimal values for all parameter placeholders in the expressions. We use the ``shifted limited-memory variable-metric method'' with rank-1 update (VAR1)~\cite{Vlcek2006} as implemented in \emph{NLOpt}\footnote{\url{https://nlopt.readthedocs.io/en/latest/}}, because we found that it reached solutions of the same quality but was slightly faster than limited-memory BFGS~\cite{fletcher2000practical} for the RAR dataset described below. ESR performs up to 340 random restarts ($\sim \text{unif}(-3, 3)$) to increase the chance of finding a global optimum. The optimizer stops only after convergence (relative and absolute tolerance $<10^{-8}$) or when reaching a maximum runtime of five minutes per restart. Therefore, parameter optimization requires most of the computational effort of ESR. For the RAR dataset, this step took approximately 5 days when distributed across 320 cores on our HPC cluster\footnote{4x Intel Xeon Phi 7210 nodes with 64 cores + 1x AMD EPYC 7713 with 64 cores}.

While ESR was designed as an exhaustive method, we can also iterate through unique expressions in random order, fit them to the dataset, and stop at any time, returning the best solution found so far. We refer to this algorithm as \emph{random search} (RS) and use it as a benchmark to quantify the success probability and efficiency of GP. Note, however, that this is an idealised random search in a much smaller search space, because it visits only unique expressions.

\subsection{Genetic programming}

We adapted the TinyGP~\cite{Sipper2019tinyGP} implementation by Moshe Sipper, a Koza-style genetic programming system for symbolic regression~\cite{koza1992} that is easy to understand and can be easily extended to include parameter optimization for non-Gaussian likelihoods, such as the MNR likelihood described below. TinyGP implements a ramped half-and-half initialization and follows the usual GP process. For a population of size $P$, $P$ new individuals are generated by: 1) selecting two parents using tournament selection; 2) applying crossover between these parents with probability $p_{cx}$, otherwise returning the first parent; 3) applying subtree mutation at a random node with probability $p_{mut}$; 4) replacing the current population with the newly generated solutions; 5) replacing the worst solution in the population with the best of all time (elitism).

In ramped half-and-half, half of the population is generated using the grow method and half using the full method. 
The initialization depth is varied between $3$ and the maximum-depth hyperparameter.
For tournament selection, $n$ solution candidates are sampled from the population and the fittest one is chosen. If there are multiple solution candidates with the best fitness value, we pick one at random.

After selecting two parents, the algorithm chooses whether to apply crossover with probability $p_{cx}$. If it does not, it simply returns the first parent. Otherwise, it picks a random node from the first parent and replaces it with a random subtree from the second parent.

The mutation operator traverses the tree returned by the crossover operator and decides whether to apply the operator at that point with probability $p_{mut}$. Once it picks a mutation point, it replaces that node with a random subtree generated using the grow method with a maximum depth of $2$.

Finally, the current population is replaced by the new generated solutions and the worst generated solution is replaced by the best solution found so far (elitism).

We have adapted the TinyGP code to support univariate functions (inv, powabs), multiple variables, parameter nodes ($\theta$), and parameter optimization using SciPy's \texttt{minimize}\footnote{https://docs.scipy.org/doc/scipy/reference/generated/scipy.optimize.minimize.html} with the default BFGS optimizer. The partial derivatives of expressions with respect to $\theta$ (and to $x$ for the MNR likelihood) are calculated via forward-mode automatic differentiation (cf.~\cite{margossian2019review}).

Additionally, for the initial population, we ensure that no expression throws an evaluation error (i.e., Inf or NaN) by resampling until it has a finite fitness value. Whenever we generate an expression that exceeds the length limit, we assign it a poor fitness value so it is discarded.

The hyperparameter values used in this work are listed in Table~\ref{tab:hyperparams}.
We ran three experiments for each dataset while varying the maximum length limit, using a larger population size for length 20. We collected logs of all evaluated expressions from 50 independent runs for both datasets and used them for our analysis.\footnote{These experiments were executed on a Dell G5 with a Core i7-9750H and 16GB of RAM.}

\begin{table}[t!]
\setlength{\tabcolsep}{3pt} 
    \centering
     \caption{List of the hyperparameters used in the experiments.}
    \begin{tabular}{l|c|c|c}
    \hline\hline
    \textbf{Parameter} & \multicolumn{3}{p{6cm}}{\centering \textbf{Value}} \\
       \hline\hline
       Max. length & $10$ & $12$ & $20$ \\
       \hline
       Pop. size & $100$ & $100$ & $500$ \\
       Generations & $250$ & $250$ & $250$ \\
       Min. depth & $2$ & $2$ & $2$ \\
       Max. depth & $4$ & $4$ & $4$ \\
       Tournament size & $2$ & $2$ & $4$ \\
       Cx. prob. & $1.0$ & $1.0$ & $1.0$ \\
       Mut. prob. & $25~\%$ & $25~\%$ & $25~\%$ \\
       Objective (Nikuradse) & min. MSE & min. MSE & min. MSE  \\
       Objective (RAR) & max. MNR logL & max. MNR logL & max. MNR logL  \\
       Optim. iterations (L-BFGS) & $10$ & $10$ & $10$ \\
       Function set & \multicolumn{3}{c}{$+, -, \div, \times, \operatorname{powabs}, \operatorname{inv}$}\\
       \hline\hline
    \end{tabular}
    \label{tab:hyperparams}
    \vspace{-4mm}
\end{table}

\subsection{Datasets}
\subsubsection{Flow in rough pipes -- Nikuradse}

We extracted the Nikuradse rough-pipe flow dataset from~\cite{nikuradse1950}. Symbolic regression results for this dataset, predicting the relationship between turbulent friction $\lambda$ and scaled roughness, have been reported in~\cite{Guimera2020,Reichardt2020}. Reichardt et al.~\cite{Reichardt2020} stressed that ``Over eight decades later, and despite the fundamental and practical importance of the problem'' the functional dependence of friction on the Reynolds number and relative roughness is still unknown. They found that their system, called \emph{Bayesian machine scientist}, prefers Prandtl's collapse over models proposed later. We use ESR and GP to find an expression for the collapse $y=f(x)$ using the scaled variables $y=\lambda^{-1/2} - 2\,\log\frac{r}{k}$ and $x=\log\frac{v_*\,k}{\nu}$, as reported in~\cite{nikuradse1950}.

\subsubsection{Radial acceleration relation (RAR)}

We also explore a real-world dataset from the field of galactic astrophysics. The Radial Acceleration Relation (RAR) describes the link between the acceleration sourced by visible stars and gas (``baryons'' in astrophysicists' jargon), $g_\text{bar}$, and the total dynamical acceleration as traced by the trajectories of dynamical tracers in galaxies, $g_\text{obs}$~\cite{RAR}. The relation is a key, although somewhat contentious, piece of evidence in the debate over the meaning of the ``missing mass problem,'' namely that galaxy dynamics appears to require significantly more mass than is visible (e.g.~\cite{McGaugh_tale}). 
In particular, the tightness of the relation (it has small intrinsic scatter, $\sigma_{\rm int}$) is often used as evidence that in fact it is not mass that is missing (the ``dark matter'' of the standard cosmological model), but rather that gravity behaves differently on galactic scales than in the Solar System. Modified Newtonian Dynamics (MOND) accommodates the relation naturally, and in fact predicted it at the theory's inception~\cite{Milgrom_1,Milgrom_2}.

The RAR has been measured in a range of galaxy types, as well as galaxy groups~\cite{Freundlich,McGaugh_Wolf,McGaugh_Milgrom,ellipticals_1,ellipticals_2,groups,Oman}. We use here the data from the SPARC sample, the flagship dataset for precision analysis of the RAR~\cite{SPARC}. This is a compilation of 175 late-type galaxies with resolved HI and H$\alpha$ rotation curves from the literature and \textit{Spitzer} 3.6 $\mu$m photometry to determine baryonic mass distributions. We remove galaxies with poor measurement quality (flag 3), galaxies with inclination $<30^\circ$, and velocity measurements with $>10\%$ uncertainty, leaving 2696 points from 147 galaxies. This follows the analysis of~\cite{ESR_RAR} (in turn following~\cite{RAR}), who were the first to apply symbolic regression---in particular ESR---to galaxy dynamics, finding many new functions that fit those data better than MOND functions.

We implement one methodological improvement over~\cite{ESR_RAR}. That study used a likelihood function that integrated over the true values of the independent variables ($g_\text{bar}$) in the dataset using a uniform prior. However, it was shown in~\cite{MNR} that this can lead to significant bias in the best-fit parameter values; these biases are removed by instead marginalising over the true $g_\text{bar}$ using a Gaussian hyperprior whose mean, $\mu$, and width, $\omega$, are free parameters to be inferred simultaneously (themselves with uniform priors) with those of the function being fitted. In particular, given a function $y = f \left( x, \bm{\theta}\right)$ with free parameters $\bm{\theta}$, the log-likelihood of data $\mathcal{D}$ is 
\begin{equation}\label{eq:MNR}
    \begin{split}
        \log \mathcal{L}(\mathcal{D}|\bm{\theta},\mu,\omega,\sigma_\text{int})
    & = - \frac{1}{2} \sum_i \frac{ \omega^2 \left( \mathcal{A}_i x_i + \mathcal{B}_i - y_i \right)^2 + {\sigma^2_{x_i}} \left( \mathcal{A}_i \mu + \mathcal{B}_i - y_i \right)^2}{\mathcal{A}_i^2 \omega^2 {\sigma^2_{x_i}} +  \left( {\sigma^2_{y_i}} + \sigma_{\rm int}^2 \right) \left( \omega^2 + {\sigma^2_{x_i}} \right)} \\
    & - \frac{1}{2} \sum_i \frac{ \left( {\sigma^2_{y_i}} + \sigma_{\rm int}^2 \right) \left( x_i - \mu \right)^2}{\mathcal{A}_i^2 \omega^2 {\sigma^2_{x_i}} + \left( {\sigma^2_{y_i}} + \sigma_{\rm int}^2 \right) \left( \omega^2 + {\sigma^2_{x_i}} \right)} \\
    & - \frac{1}{2} \sum_i \log \left(\mathcal{A}_i^2 \omega^2 {\sigma^2_{x_i}} +\left( {\sigma^2_{y_i}} + \sigma_{\rm int}^2 \right) \left( \omega^2 +{\sigma^2_{x_i}} \right) \right)+ {\rm const.},
    \end{split}
\end{equation}
where
\begin{equation}
    \mathcal{A}_i \equiv  \left. \frac{\partial}{\partial x} f \left( x, \bm{\theta}\right) \right|_{x = x_i}, \quad 
    \mathcal{B}_i \equiv f \left( x_i, \bm{\theta}\right) - \mathcal{A}_i x_i.
\end{equation}
$i$ indexes the data points and
$\sigma_{x_i}$ and $\sigma_{y_i}$ are the uncertainties on the $x$ and $y$ variables. For RAR, $x$ is $g_\text{bar}$ and $y$ is $g_\text{obs}$.

We have verified separately to~\cite{MNR} that MNR leads to unbiased parameter recovery on mock RAR-like datasets, while the uniform prior method used in~\cite{ESR_RAR} does not. 
We employ MNR here by maximising Eq.~\ref{eq:MNR} simultaneously in $\bm{\theta},\mu,\omega$ and $\sigma_\text{int}$.

\section{Results}
\label{sec:results}

\subsection{Characterization of the solution space}

\begin{figure}[t!]
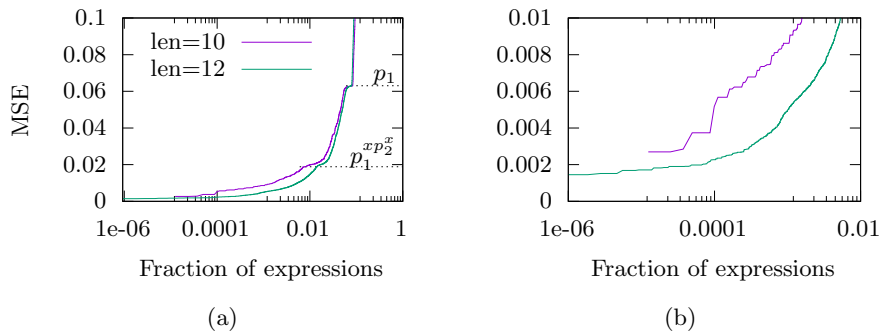

    \centering
    \begin{subfigure}[b]{0.49\textwidth}
    \input{fig/nikuradse_2_distr.tex}
    \caption{}
    \label{fig:nikuradse-2-distribution}
    \end{subfigure}
    \begin{subfigure}[b]{0.49\textwidth}
    \input{fig/nikuradse_2_distr_zoom.tex}
    \caption{}
    \label{fig:nikuradse-2-distribution-zoom}
    \end{subfigure}    
    \caption{Distribution of MSE values for all possible expressions with the Nikuradse  dataset (a) and a zoomed region (b). The constant model $p_1$ has an MSE $=0.063$ and only 10~\% of the solutions have a better MSE. The expression $p_1 ^{x p_2^x}$ reaches MSE=0.019 which only around 1~\% of the expressions surpass. The subplot on the right hand side shows that MSE less than 0.002 is reached only by about the 100 best expressions with length 12.}
    \label{fig:nikuradse-2-distribution-full}
\end{figure}

Figures~\ref{fig:nikuradse-2-distribution} and \ref{fig:nikuradse-2-distribution-zoom} show the distribution of MSE values for unique expressions on the Nikuradse dataset with maximum lengths $10$ and $12$. The best expression for length $10$ has MSE $=2.7\cdot 10^{-3}$, and the best expression for length $12$ has MSE $=1.46\cdot 10^{-3}$. Only about $0.01~\%$ of all expressions have an MSE close to the minimum; most expressions have an MSE about $10\times$ higher. It is also worth noting that only $10~\%$ of expressions are better than the baseline model $p_1$, and only $1~\%$ are better than $p_1^{x\, p_2^x}$.

We can see the top-$5$ expressions for size $10$ and $12$ in Table~\ref{tab:niku-2-top-5-expressions}. The top expressions share some similarities in their form as the adjustable parameters can compensate for any smaller differences. The list shows semantic duplicates with equivalent likelihood values that have not been detected as a consequence of the limited number of eq-sat iterations or missing rules. 

\begin{table}[t!]
\caption{Top-$5$ expressions of size $12$ and $10$ for the Nikuradse dataset}
\scriptsize
\begin{tabular}{clclc}
\hline\hline
Max. len. & Expression & LogLik & Parameters ($[p_1 \ldots p_5]$) & MSE \\
\hline
12 & ${p_1} / (1.0 / ({p_2} - x) - {p_3} ^ {{p_4} ^ x})$ & 668.65 & $[-2.222, -0.253, 3.433\cdot 10^{-5}, 0.137]$ & $1.456 \cdot 10^{-3}$\\
12 & ${p_1} / (1.0 / ({p_2} + x) + {p_3} ^ {{p_4} ^ x})$ & 668.65 & $[2.222, 0.253, -3.433\cdot 10^{-5}, -0.137]$ &$ 1.456 \cdot 10^{-3}$ \\
12 & ${p_1} / (|x| ^ {-x} - {p_2} ^ {{p_3} ^ x})$ & 661.19 & $[-1.928, -4.043, 0.463]$ & $1.517 \cdot 10^{-3}$ \\
12 & ${p_1} / ({p_2} ^ {{p_3} ^ x} - |1.0 / x| ^ x)$ & 661.19 & $[1.928, 4.043, 0.463]$ & $1.517 \cdot 10^{-3}$ \\
12 & ${p_1} / (|1.0 / x| ^ x - {p_2} ^ {{p_3} ^ x})$ & 661.20 & $[-1.928, -4.043, -0.463]$ & $1.517 \cdot 10^{-3}$ \\
\hline
10 & ${p_1} / (1.0 / ({p_2} + x) - {p_3} ^ x)$ & 556.94 & $[0.301, 0.673, -0.453]$ &$ 2.699\cdot 10^{-3}$ \\
10 & ${p_1} / (1.0 / ({p_2} - x) + {p_3} ^ x)$ & 556.94 & $[-0.301, -0.673, 0.453]$ &$ 2.699\cdot 10^{-3}$ \\
10 & $1.0 / ({p_1} + |{p_2} + {p_3} ^ x| ^ {p_4})$ & 547.04 & $[0.456, -0.191, -0.177, 1.216]$ &$ 2.851\cdot 10^{-3}$ \\
10 & ${p_1} - |{p_2} + -1.0 / ({p_3} + x)| ^ {p_4}$ & 497.93 & $[2.262, 0.773, 0.301, 0.677]$ & $3.739 \cdot 10^{-3}$\\
10 & ${p_1} - |{p_2} + 1.0 / ({p_3} + x)| ^ {p_4}$ & 497.93 & $[2.262, -0.7735, 0.301, 0.6767]$ & $3.739\cdot 10^{-3}$ \\
\hline\hline
\end{tabular}
\label{tab:niku-2-top-5-expressions}
\end{table}

\begin{figure}[t!]
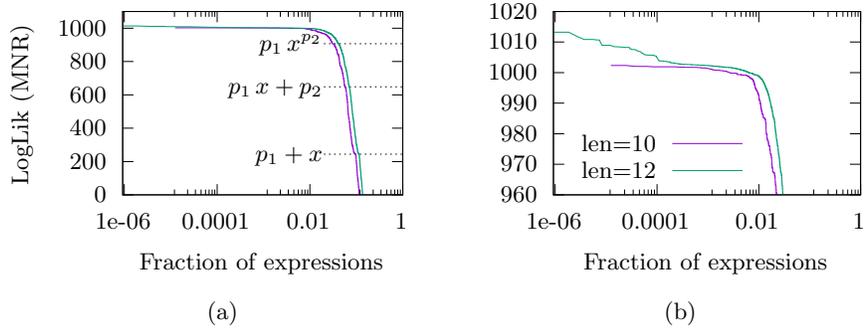

    \centering
    \begin{subfigure}[b]{0.49\textwidth}
    \input{fig/rar_distr.tex}
    \caption{}
    \label{fig:rar-distribution}
    \end{subfigure}
    \begin{subfigure}[b]{0.49\textwidth}
    \input{fig/rar_distr_zoom.tex}
    \caption{}
    \label{fig:rar-distribution-zoom}
    \end{subfigure}    
    \caption{Distribution of log-likelihood values for all possible expressions for the RAR dataset (a) and a zoomed region (b). Around 10~\% of all solutions reach a good log-likelihood $\approx 1000$. The zoomed plot shows that approximately only the 100 best solutions with length 12 have a log-likelihood above 1005.}
    \label{fig:rar-distribution-FULL}
\end{figure}

Figures~\ref{fig:rar-distribution} and \ref{fig:rar-distribution-zoom} show the distribution of MNR log-likelihood values for the RAR dataset for the unique expressions with maximum length 10 and 12. The distribution shows a large subset of solutions (approximately $1~\%$ of all solutions) with a log-likelihood around $1000$. The best solution for length $10$ has log-likelihood $1002.34$, the best for length $12$ has log-likelihood $1013.24$.

The top-$5$ solutions for the RAR dataset are listed in Table~\ref{tab:rar-top-5-expressions}. For these expressions, the MNR likelihood parameters are consistently estimated as $\sigma_\text{int}\approx 0.08, \mu \approx -0.50, \omega \approx 0.74$.

\begin{table}[t!]
\scriptsize
\centering
\caption{Top-$5$ expressions for the RAR dataset}
\begin{tabular}{clcl}
\hline\hline
Max. len. & Expression & NegLogLik & Parameters ($[p_1 \ldots p_4, \sigma_{\text{int}}, \mu, \omega]$)\\
\hline
12 & ${p_1}\, x ^ {-1.0 / ({p_2} + |{p_3} + x| ^ {p_4})}$ & -1013.24 & $[-1.73, -2.41, -0.0179, 0.0513, 0.0792, -0.502, 0.74]$\\
12 & ${p_1}\, x ^ {1.0 / ({p_2} - |{p_3} + x| ^ {p_4})}$ & -1013.24 & $[-1.73, 2.41, -0.0179, 0.0513, -0.0792, -0.502, 0.74]$\\
12 & $1.0 / ({p_1} ^ {|{p_2} + x| ^ {p_3}} - x ^ {p_4})$ & -1011.76 & $[0.407, -0.0175, 0.221, -0.502, 0.0796, -0.502, 0.74]$\\
12 & ${p_1} + |{p_2} + -1.0 / ({p_3} - x)| ^ x - x$ & -1010.84 & $[-1.025, -0.775, -1.272, -0.0791, -0.503, 0.74]$\\
12 & ${p_1} + |{p_2} + 1.0 / ({p_3} - x)| ^ x - x$ & -1010.84 & $[-1.025, 0.775, -1.272, -0.0791, -0.503, 0.74]$\\
\hline
10 & $1.0 / ({p_1} + |{p_2} + x ^ {{p_3}}| ^ {{p_4}})$ & -1002.34 & $[0.018, -0.276, -0.297, 1.752, 0.0806, -0.503, 0.74]$ \\
10 & $1.0 / ({p_1} - |{p_2} + x ^ {p_3}| ^ {p_4})$ & -1002.34 & $[-0.018, -0.276, 0.297, 1.752, 0.0806, -0.503, 0.74]$ \\
10 & ${p_1} ^ {{p_2} - x ^ {p_3}} - x$ & -1002.06 & $[1.39 \cdot 10^{-5}, 0.911, 0.0723, 0.0806, -0.503, 0.74]$ \\
10 & ${p_1} ^ {{p_2} - (1.0 / x) ^ {p_3}} - x$ & -1002.06 & $[-1.39 \cdot 10^{-5}, 0.911, -0.0722, 0.0806, -0.503, 0.74]$ \\
10 & $x - {p_1} ^ {{p_2} - x ^ {p_3}}$ & -1002.06 & $[1.39 \cdot 10^{-5}, 0.911, 0.0722, 0.0806, -0.503, 0.74]$ \\
\hline\hline
\end{tabular}
\label{tab:rar-top-5-expressions}
\end{table}

\subsection{Quality of GP solutions}

Table~\ref{tab:gp-results} shows the results obtained by GP with the different maximum length limits for each dataset. We report the objective values of the best solution found over 50 independent runs, as well as the average and standard deviation of the objective values of each of the 50 solutions. For both datasets and both length limits, GP was not able to find the global optimum. Only when allowing longer expressions, GP was capable of finding a better (but longer) solution for the Nikuradse dataset.
Looking at the average objective, we see that only the larger length limit yields solutions whose objective values are consistently closer to the best solutions from the more restricted search spaces.

To verify that our TinyGP implementation is representative of other GP implementations, we also ran experiments on the Nikuradse dataset with Operon, a state-of-the-art GP implementation for SR, using the same settings. Since Operon only supports optimization of the sum of squares, we could not run the RAR experiments with it. The Operon results in Table~\ref{tab:gp-results} are slightly better but qualitatively similar to the TinyGP results.

\begin{table}
\centering
\caption{Objective values of the best solutions found by GP compared to the global optimum in the search space found by ESR. Smaller values are better. For the Nikuradse dataset the MSE is reported, for the RAR dataset the negative MNR log-likelihood. For the Nikuradse dataset the results for Operon are similar to the results achieved with TinyGP. Neither TinyGP nor Operon find the global optimum}

\setlength{\tabcolsep}{6pt} 
\begin{tabular}{lcrrrr}

\hline\hline
Dataset & Max. length & Optimum          & Best           & Mean          & StdDev \\
\hline
Nikuradse & 10   & $2.70\cdot 10^{-3}$ & $ 4.80 \cdot 10^{-3}$ & $7.56 \cdot 10^{-3}$ & $2.97\cdot 10^{-3}$\\
Nikuradse & 12   & $1.46\cdot 10^{-3}$ & $ 2.01 \cdot 10^{-3}$ & $5.39 \cdot 10^{-3}$ & $1.49\cdot 10^{-3}$\\
Nikuradse & 20   &                     & $ 1.30 \cdot 10^{-3}$ & $1.33 \cdot 10^{-3}$ & $0.03\cdot 10^{-3}$\\
RAR         & 10   &  $-1002.34$         & $ -1000.66$           & $ -999.23$           & $ 0.225$ \\ 
RAR         & 12   &  $-1013.24$         & $ -1001.65$           & $ -999.44$           & $ 0.630$ \\
RAR         & 20   &                     & $ -1007.08$           & $-1002.31$           & $ 0.934$ \\
\hline\hline
\end{tabular}

\begin{tabular}{lccccc}
Nikuradse (Operon) & 10 & $2.70\cdot 10^{-3}$ & $2.86 \cdot 10^{-3}$ & $8.57 \cdot 10^{-3}$ & $2.86 \cdot 10^{-3}$ \\
Nikuradse (Operon) & 12 & $1.46\cdot 10^{-3}$ & $1.68 \cdot 10^{-3}$ & $2.51 \cdot 10^{-3}$ & $0.58 \cdot 10^{-3}$ \\ 
Nikuradse (Operon) & 20 & & $1.24 \cdot 10^{-3}$ & $1.28 \cdot 10^{-3}$ & $0.02 \cdot 10^{-3}$ \\
\hline\hline
\end{tabular}

\label{tab:gp-results}

\end{table}

\subsection{Success probability}
To better understand why GP does not find the global optimum on either dataset, we analyse empirical cumulative density functions (ECDFs) of the number of visited expressions required to find a solution (cf.~\cite{hansen2021coco}). To compare search efficiency, we plot multiple ECDF curves in one figure for different objective thresholds and algorithms. The reported success probability is the fraction of the 50 independent runs in which a threshold is reached, and the x-axis shows the number of expressions visited before that threshold is reached.

We use ESR-based RS as an (idealized) benchmark that ignores the effort for parameter optimization and the problem of local optima, and allows us to focus on the search efficiency of expression trees. 

\begin{figure}[t!]
    \centering
    \input{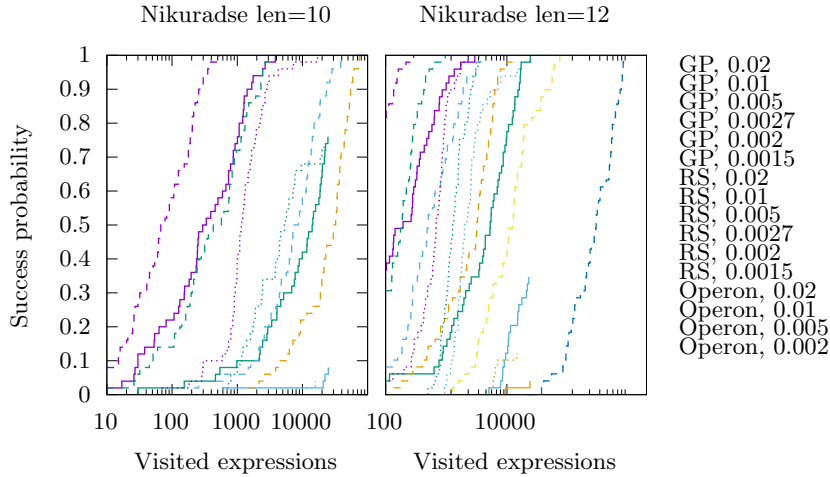}
    \caption{Success probability of GP and RS over the number of visited expressions for length=10 and length=12 on the Nikuradse dataset. For length=10, GP has a high probability of finding solutions with MSE below 0.02 and 0.01, but its success rate drops below 10~\% for a threshold of 0.005. For length=12, success rates are higher, but GP did not find the best solutions in any of the 50 runs. Operon performs better than TinyGP for length=12 but is still slower than RS.}
    
    \label{fig:success-probability-nikuradse2}
\end{figure}

In Fig.~\ref{fig:success-probability-nikuradse2}, the ECDF curves for multiple MSE thresholds on the Nikuradse dataset show that RS has a higher success probability and requires fewer visited expressions for all thresholds. RS also always finds the best solution in the finite solution space because it does not resample solutions.
The first threshold of 0.02 is deliberately high, since approximately 10~\% of solutions have a better MSE. For length$=10$, GP needs about 600 visited expressions to reach 50~\% success and reaches 100~\% success after about 4000 visited expressions. RS reaches 50~\% success after about 100 visited expressions and 100~\% success after fewer than $1000$. For lower thresholds, both algorithms need more time, and GP success probability decreases because we limit each GP run to at most 25000 expressions (population size 100, 250 generations). GP finds no solutions below thresholds $0.002$ and $0.0015$. 
This pattern is repeated for maximum length $12$.

\begin{figure}[t!]
    \centering
    \input{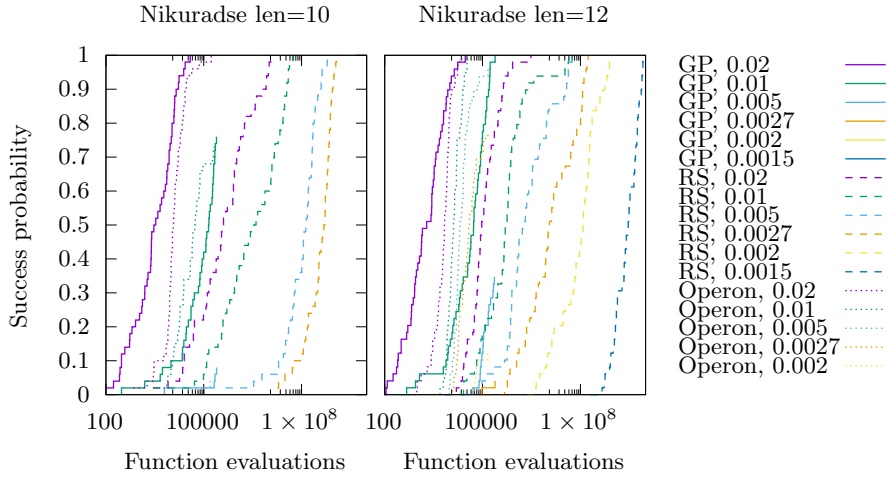}
    \caption{Success probability of GP and RS over the number of function evaluations for length=10 and length=12 for the Nikuradse dataset. The success probabilities are the same as in Figure~\ref{fig:success-probability-nikuradse2} but GP is more efficient than RS when counting the number of function evaluations.}
    \label{fig:success-probability-nikuradse2-fevals}
\end{figure}

Instead of plotting success probability over the number of visited expressions, one could make a more realistic comparison that includes parameter-optimization effort in ESR by plotting success over the number of function evaluations, as in Figure~\ref{fig:success-probability-nikuradse2-fevals}. This plot clearly shows that GP is much faster, because it performs only a single BFGS run with 10 iterations for each visited expression. We deliberately ignore this optimization effort here, since our focus is the efficiency of GP in sampling expression structures. 

\begin{figure}[t!]
    \centering
    \input{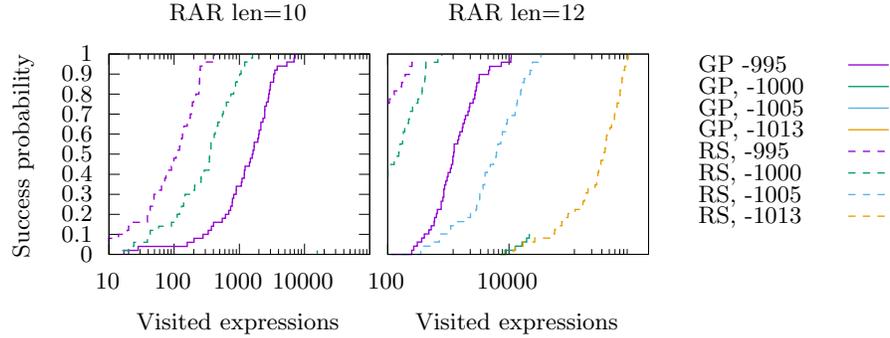}
    \caption{Success probability over number of visited expressions for RAR for length=10 and length=12. For length=10, GP finds solutions with nll below $-995$ easily but found only a single solution with nll $< -1000$ in 50 runs. For length=12, GP has a  success rate of $\approx 15~\%$ for nll $< -1000$, but fails to find one of the $\approx 100$ best solutions with nll $<-1005$.}
    \label{fig:success-probability-rar}
\end{figure}

Fig.~\ref{fig:success-probability-rar} shows the ECDF for RS and GP for the RAR dataset with a similar result. GP does not find the best solutions, and has to visit more expressions than RS until it finds solutions. The success rate of reaching a solution with neg. log-likelihood (nll) below $-1000$ is only approximately $15~\%$ even for maximum length 12. 

Given these results, the question remains whether expanding the GP search space enables it to find expressions with the same MSE and length as RS. Fig.~\ref{fig:success-probability-mixed} compares RS with maximum length $12$ against GP with maximum length $20$ (restricted to solutions of the same length as RS). Surprisingly, RS is still more efficient: it reaches 100\% success for MSE below $0.002$, whereas GP success is capped at 50\%.

\subsection{Semantic duplicates}

The previous plots highlight an issue with the power of GP when compared to RS. 
One reason is wasteful exploration of repeated expressions. GP can revisit expressions in two situations: expressions may be regenerated by chance through crossover and mutation, or GP may generate a structurally different expression that is equivalent after simplification. The same eq-sat-based simplification used in ESR to produce semantically distinct expressions can also be used to analyse how many GP-evaluated expressions are duplicates.

\begin{figure}[t!]
    \centering
    \input{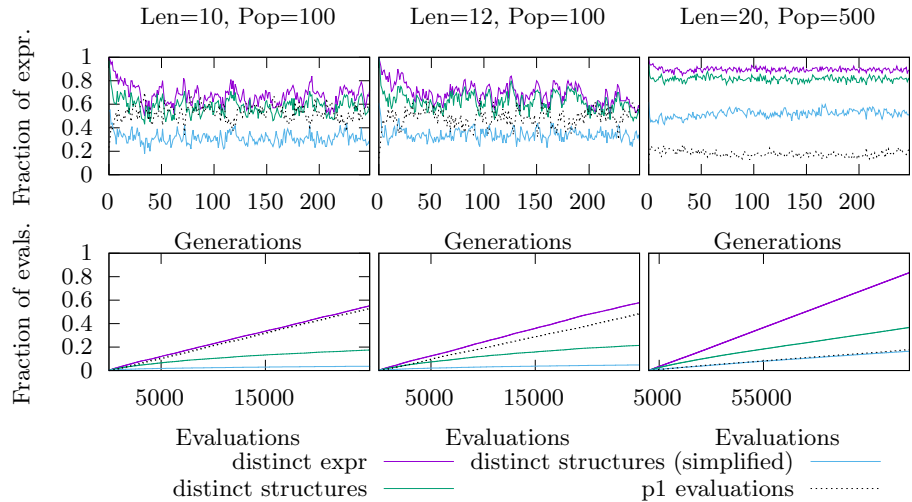}
    \caption{Number of expressions, and distinct expression structures visited by a single GP run for the Nikuradse dataset relative to population size (top row) and number of total evaluations (bottom row).}
    \label{fig:visited-expressions-tinygp-nikuradse2}
\end{figure}

\begin{figure}[t!]
    \centering
    \input{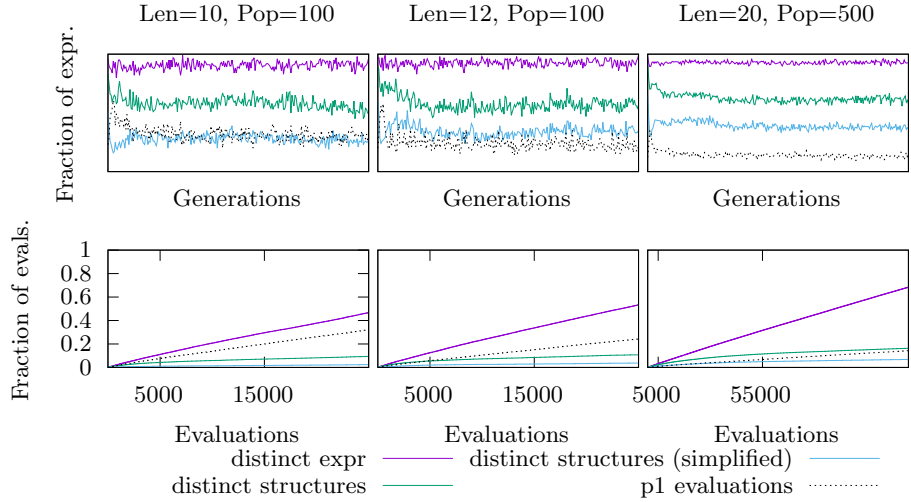}
    \caption{Number of expressions, and distinct expression structures visited by Operon for the Nikuradse dataset, relative to population size (top row) and number of total evaluations (bottom row).}
    \label{fig:visited-expressions-operon-nikuradse2}
\end{figure}

Fig.~\ref{fig:visited-expressions-tinygp-nikuradse2} shows the number of distinct expressions visited throughout the search for the three length limits. The top row of the plot shows the number of distinct expressions, distinct expression structures (ignoring parameter values), distinct expression structures after simplification, and expressions that can be simplified to the trivial expression $p_1$ (a constant model) for each generation. The frequencies are given as fractions of the population size. The plots in the bottom row show the aggregates over the whole run, where distinctness is determined over the whole run.
These plots reveal a worrying behaviour of GP when restricted to such extreme length limits. Without applying simplification, GP generates between $60$ to $80~\%$ of distinct expressions ($20~\%$ of the expressions or expression structures are the same). When we simplify such expressions, this range goes down to only $20$ to $40~\%$. This behaviour is similar for maximum lengths $10$ and $12$. For maximum length $20$, the search is a little less wasteful, with distinct expressions after simplification in the range of $40\%$ and $60\%$.
Even more alarming is that around 50~\% of the expressions can be simplified to a constant $(p_1)$ for length=10 and length=12, and for length=20 this is still $10$ to $20~\%$. The accumulated counts in the bottom row show that only $10$ to $20~\%$ of all expressions that GP visits are unique.
Notice that the expression $p_1$ can be generated in many forms, such as $p_1 + p_2$ or $p_1 / p_2^{p_3}$. The high percentage of such trivial expression being sampled by GP is related to its position in the search space in which it has a reasonable MSE and many different expressions with MSE worse than this one.
As shown in Figure~\ref{fig:visited-expressions-operon-nikuradse2}, the results for Operon are qualitatively similar to the TinyGP results.

\begin{figure}[t!]
    \centering
     \input{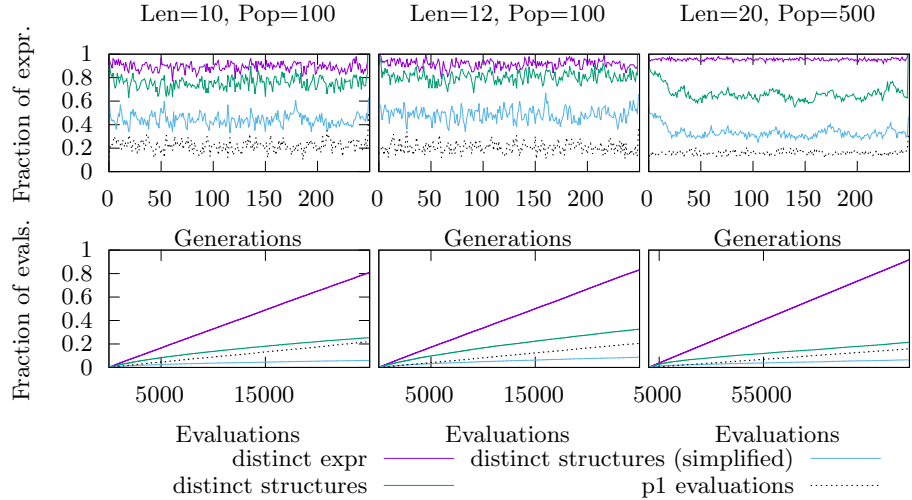}
    \caption{Number of expressions and distinct expression structures visited by a single GP run for RAR relative to population size (top row) and number of total evaluations (bottom row).}
    \label{fig:visited-expressions-tinygp-rar}
\end{figure}

We observe similar behaviour for the RAR dataset, as shown in Fig.~\ref{fig:visited-expressions-tinygp-rar}. The main difference is that GP visits more distinct expressions and structures ($80$ to $100~\%$), possibly because a large fraction of expressions have likelihood values close to the best one. However, the number of semantically distinct expressions after simplification is again much lower, around $50~\%$. A main contributor is again expressions that simplify to a constant $(p_1)$, which account for about $20~\%$ of expressions. The aggregated counts of distinct expressions are similarly low as for Nikuradse: only about $10~\%$ of evaluated expressions are unique, so about $90~\%$ are revisits. 

\section{Limitations}
\label{sec:limits}

We stress that these experiments investigate visited expressions only, without considering lineage (i.e., the exploration path used to reach an optimal expression). We focus on a broader view of GP efficiency as a search algorithm for SR by comparing it with the enumerated search space and an idealised random search that samples without repetition. 

We used an adapted version of TinyGP, which is much simpler than state-of-the-art GP systems such as Operon~\cite{burlacu2020operon} or PySR~\cite{cranmerpysr}. To ensure that the results are not an artefact of TinyGP, we also ran the same experiments on the Nikuradse dataset with Operon and found similar results. The number of unique expressions visited by Operon is similarly low. 

We restricted the GP expressions to short length limits. This was necessary to have the same search space for ESR and GP. While we did quantify the efficiency of GP for a larger length limit of 20 nodes, we cannot calculate the success probability for finding the best solutions in the larger search spaces. 

\begin{figure}[t!]
    \centering
    \input{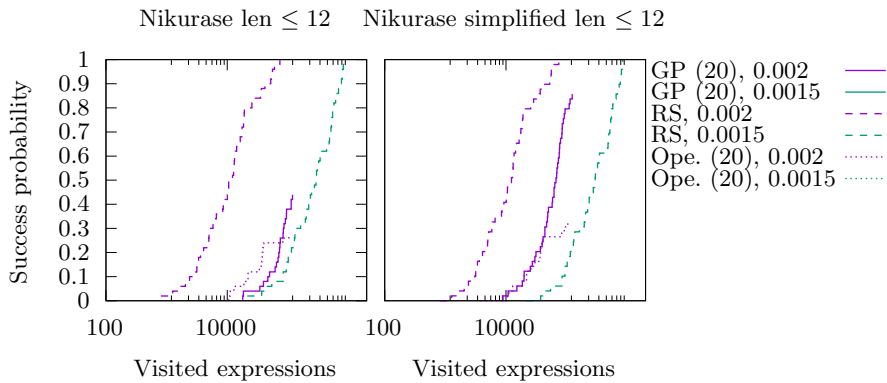}
    \caption{Success probability over number of visited expressions for RS in the length=12 search space compared to GP with a larger length limit (=20). Here, success is defined as finding a solution with MSE below the threshold \emph{and} length less than or equal to 12.}
    \label{fig:success-probability-mixed}
\end{figure}

GP may require larger length limits to find the best solutions, which can then be simplified. We analysed GP runs with length limit 20 and plotted ECDFs for the number of visited expressions needed to find solutions whose simplified length is at most 12 nodes. 
As shown in Figure~\ref{fig:success-probability-mixed}, both TinyGP and Operon can then find expressions with MSE below 0.002 at a high success rate (compare Fig.~\ref{fig:success-probability-nikuradse2}), but they still fail to find the global optimum. Both GP systems produce similar results.
We leave a more detailed study of this effect for future work.

\section{Discussion and conclusions}
\label{sec:conclusion}

We investigated the search behaviour of genetic programming for symbolic regression by applying a simple implementation called TinyGP to two real-world datasets from the physical sciences. We focused on symbolic regression of functions of a single variable with short length limits (10 -- 20 nodes). For this search space we were able to enumerate the space of all semantically unique expressions using an improved implementation of the Exhaustive Symbolic Regression algorithm to verify how much of this space is explored by GP and the extent to which this exploration is efficient.

Having the enumerated solution space, we could first verify that in both datasets a large portion of the expressions are of low quality and only $0.01$ to $1~\%$ (depending on the dataset) are within a reasonable distance from the optimum in the fitness space. 
Using 50 independent restarts of GP we did not find the optimal solutions for any of the search spaces. 
Comparing GP success rates with an idealised random search, we found that GP visits roughly an order of magnitude more expressions to achieve the same success rate.

Finally, we showed that a significant portion of expressions visited by GP are semantically equivalent to previously visited expressions. We determine semantic equivalence by simplifying expressions to a canonical form using equality saturation, which allows detection of isomorphic expressions even when parameter values differ. This is an important novelty compared with existing SR literature, where semantic equivalence is usually detected only through shared fitness values.
From our results, it remains unclear whether the observed number of reevaluations is helpful for GP, or whether preventing them would improve search efficiency. Although representation redundancy grows exponentially with search-space size, it is unknown whether the same proportion of revisits occurs when GP uses more typical length limits (up to 100 nodes) or more input variables. We leave this topic for future work.

\section*{Acknowledgements}
We thank Pedro Ferreira for enabling this collaboration and his ideas for improving  this work, and Bogdan Burlacu for his help when making changes to Operon.
Gabriel Kronberger is supported by the Austrian Federal Ministry for Climate Action, Environment, Energy, Mobility, Innovation and Technology, the Federal Ministry for Labour and Economy, and the regional government of Upper Austria within the COMET project ProMetHeus (904919) supported by the Austrian Research Promotion Agency (FFG).
Fabricio Olivetti de Franca is supported by CNPq through the grant 301596/2022-0. 
Harry Desmond is supported by a Royal Society University Research Fellowship (grant no. 211046).
Deaglan J. Bartlett is supported by the Simons Collaboration on ``Learning the Universe.''

\bibliographystyle{splncs04}
\bibliography{symreg.bib}

@book{kronberger2024,
  title = {Symbolic Regression},
  author = {Gabriel Kronberger and Bogdan Burlacu and Michael Kommenda and Stephan M. Winkler and Michael Affenzeller},
  publisher = "Chapman \& Hall / CRC Press",
  year = "2024",  
}

@ARTICLE{McGaugh_tale,
       author = {{McGaugh}, Stacy S.},
        title = "{A tale of two paradigms: the mutual incommensurability of {\ensuremath{\Lambda}}CDM and MOND}",
      journal = {Canadian Journal of Physics},
         year = 2015,
        month = feb,
       volume = {93},
       number = {2},
        pages = {250-259},
          doi = {10.1139/cjp-2014-0203},
archivePrefix = {arXiv},
       eprint = {1404.7525},
 primaryClass = {astro-ph.CO},
       adsurl = {https://ui.adsabs.harvard.edu/abs/2015CaJPh..93..250M}
}

@ARTICLE{RAR,
       author = {{Lelli}, Federico and {McGaugh}, Stacy S. and {Schombert}, James M. and {Pawlowski}, Marcel S.},
        title = "{One Law to Rule Them All: The Radial Acceleration Relation of Galaxies}",
      journal = {Astrophysical Journal},
         year = 2017,
        month = feb,
       volume = {836},
       number = {2},
          eid = {152},
        pages = {152},
          doi = {10.3847/1538-4357/836/2/152},
archivePrefix = {arXiv},
       eprint = {1610.08981},
 primaryClass = {astro-ph.GA},
       adsurl = {https://ui.adsabs.harvard.edu/abs/2017ApJ...836..152L}
}

@ARTICLE{SPARC,
   author = {{Lelli}, F. and {McGaugh}, S.~S. and {Schombert}, J.~M.},
    title = "{SPARC: Mass Models for 175 Disk Galaxies with Spitzer Photometry and Accurate Rotation Curves}",
  journal = {Astronomical Journal},
archivePrefix = "arXiv",
   eprint = {1606.09251},
     year = 2016,
    month = dec,
   volume = 152,
      eid = {157},
    pages = {157},
      doi = {10.3847/0004-6256/152/6/157},
   adsurl = {http://adsabs.harvard.edu/abs/2016AJ....152..157L}
}

@ARTICLE{Milgrom_1,
   author = {{Milgrom}, M.},
    title = "{A modification of the Newtonian dynamics as a possible alternative to the hidden mass hypothesis}",
  journal = {Astrophysical Journal},
     year = 1983,
    month = jul,
   volume = 270,
    pages = {365-370},
      doi = {10.1086/161130},
   adsurl = {http://adsabs.harvard.edu/abs/1983ApJ...270..365M}
}

@ARTICLE{Milgrom_2,
   author = {{Milgrom}, M.},
    title = "{A Modification of the Newtonian Dynamics - Implications for Galaxy Systems}",
  journal = {Astrophysical Journal},
     year = 1983,
    month = jul,
   volume = 270,
    pages = {384},
      doi = {10.1086/161132},
   adsurl = {http://adsabs.harvard.edu/abs/1983ApJ...270..384M}
}

@ARTICLE{Freundlich,
       author = {{Freundlich}, Jonathan and {Famaey}, Benoit and {Oria}, Pierre-Antoine and {B{\'\i}lek}, Michal and {M{\"u}ller}, Oliver and {Ibata}, Rodrigo},
        title = "{Probing the radial acceleration relation and the strong equivalence principle with the Coma cluster ultra-diffuse galaxies}",
      journal = {Astronomy and Astrophysics},
         year = 2022,
        month = feb,
       volume = {658},
          eid = {A26},
        pages = {A26},
          doi = {10.1051/0004-6361/202142060},
archivePrefix = {arXiv},
       eprint = {2109.04487},
 primaryClass = {astro-ph.GA},
       adsurl = {https://ui.adsabs.harvard.edu/abs/2022A&A...658A..26F}
}

@ARTICLE{McGaugh_Wolf,
       author = {{McGaugh}, Stacy S. and {Wolf}, Joe},
        title = "{Local Group Dwarf Spheroidals: Correlated Deviations from the Baryonic Tully-Fisher Relation}",
      journal = {Astrophysical Journal},
         year = 2010,
        month = oct,
       volume = {722},
       number = {1},
        pages = {248-261},
          doi = {10.1088/0004-637X/722/1/248},
archivePrefix = {arXiv},
       eprint = {1003.3448},
 primaryClass = {astro-ph.CO},
       adsurl = {https://ui.adsabs.harvard.edu/abs/2010ApJ...722..248M}
}

@ARTICLE{McGaugh_Milgrom,
       author = {{McGaugh}, Stacy and {Milgrom}, Mordehai},
        title = "{Andromeda Dwarfs in Light of MOND. II. Testing Prior Predictions}",
      journal = {Astrophysical Journal},
         year = 2013,
        month = oct,
       volume = {775},
       number = {2},
          eid = {139},
        pages = {139},
          doi = {10.1088/0004-637X/775/2/139},
archivePrefix = {arXiv},
       eprint = {1308.5894},
 primaryClass = {astro-ph.CO},
       adsurl = {https://ui.adsabs.harvard.edu/abs/2013ApJ...775..139M}
}

@ARTICLE{ellipticals_1,
       author = {{Chae}, Kyu-Hyun and {Bernardi}, Mariangela and {Dom{\'\i}nguez S{\'a}nchez}, Helena and {Sheth}, Ravi K.},
        title = "{On the Presence of a Universal Acceleration Scale in Elliptical Galaxies}",
      journal = {Astrophysical Journal, Letters},
         year = 2020,
        month = nov,
       volume = {903},
       number = {2},
          eid = {L31},
        pages = {L31},
          doi = {10.3847/2041-8213/abc2d3},
archivePrefix = {arXiv},
       eprint = {2010.10779},
 primaryClass = {astro-ph.GA},
       adsurl = {https://ui.adsabs.harvard.edu/abs/2020ApJ...903L..31C}
}

@ARTICLE{ellipticals_2,
       author = {{Chae}, Kyu-Hyun and {Bernardi}, Mariangela and {Sheth}, Ravi K. and {Gong}, In-Taek},
        title = "{Radial Acceleration Relation between Baryons and Dark or Phantom Matter in the Supercritical Acceleration Regime of Nearly Spherical Galaxies}",
      journal = {Astrophysical Journal},
         year = 2019,
        month = may,
       volume = {877},
       number = {1},
          eid = {18},
        pages = {18},
          doi = {10.3847/1538-4357/ab18f8},
archivePrefix = {arXiv},
       eprint = {1707.08280},
 primaryClass = {astro-ph.GA},
       adsurl = {https://ui.adsabs.harvard.edu/abs/2019ApJ...877...18C}
}

@ARTICLE{groups,
       author = {{Gopika}, K. and {Desai}, Shantanu},
        title = "{A test of constancy of dark matter halo surface density and radial acceleration relation in relaxed galaxy groups}",
      journal = {Physics of the Dark Universe},
         year = 2021,
        month = sep,
       volume = {33},
          eid = {100874},
        pages = {100874},
          doi = {10.1016/j.dark.2021.100874},
archivePrefix = {arXiv},
       eprint = {2106.07294},
 primaryClass = {astro-ph.CO},
       adsurl = {https://ui.adsabs.harvard.edu/abs/2021PDU....3300874G}
}

@ARTICLE{Oman,
       author = {{Oman}, Kyle A. and {Brouwer}, Margot M. and {Ludlow}, Aaron D. and {Navarro}, Julio F.},
        title = "{Observational constraints on the slope of the radial acceleration relation at low accelerations}",
      journal = {arXiv e-prints},
         year = 2020,
        month = jun,
          eid = {arXiv:2006.06700},
        pages = {arXiv:2006.06700},
archivePrefix = {arXiv},
       eprint = {2006.06700},
 primaryClass = {astro-ph.GA},
       adsurl = {https://ui.adsabs.harvard.edu/abs/2020arXiv200606700O}
}

@ARTICLE{ESR_RAR,
       author = {{Desmond}, Harry and {Bartlett}, Deaglan J. and {Ferreira}, Pedro G.},
        title = "{On the functional form of the radial acceleration relation}",
      journal = {Monthly Notices of the RAS},
         year = 2023,
        month = may,
       volume = {521},
       number = {2},
        pages = {1817-1831},
          doi = {10.1093/mnras/stad597},
archivePrefix = {arXiv},
       eprint = {2301.04368},
 primaryClass = {astro-ph.GA},
       adsurl = {https://ui.adsabs.harvard.edu/abs/2023MNRAS.521.1817D}
}

@ARTICLE{MNR,
       author = {{Bartlett}, Deaglan J. and {Desmond}, Harry},
        title = "{Marginalised Normal Regression: Unbiased curve fitting in the presence of x-errors}",
      journal = {The Open Journal of Astrophysics},
         year = 2023,
        month = nov,
       volume = {6},
          eid = {42},
        pages = {42},
          doi = {10.21105/astro.2309.00948},
archivePrefix = {arXiv},
       eprint = {2309.00948},
 primaryClass = {stat.ME},
       adsurl = {https://ui.adsabs.harvard.edu/abs/2023OJAp....6E..42B}
}

@misc{Sipper2019tinyGP,
  author = {Sipper, M.},
  title = {Tiny Genetic Programming in {P}ython},
  year = {2019},
  publisher = {GitHub},
  journal = {GitHub repository},
  howpublished = {\url{https://github.com/moshesipper/tiny_gp} }
}

@book{Koza1992,
	address = {Cambridge, MA, USA},
	author = {John R. Koza},
	isbn = {0-262-11170-5},
	publisher = {MIT Press},
	title = {Genetic Programming: On the Programming of Computers by Means of Natural Selection},
	year = {1992}
}

@article{margossian2019review,
  title={A review of automatic differentiation and its efficient implementation},
  author={Margossian, Charles C},
  journal={Wiley interdisciplinary reviews: data mining and knowledge discovery},
  volume={9},
  number={4},
  pages={e1305},
  year={2019},
  publisher={Wiley Online Library}
}

@book{norvig2002modern,
  author = {Russell, Stuart and Norvig, Peter},
  edition = 3,
  publisher = {Prentice Hall},
  title = {Artificial Intelligence: A Modern Approach},
  year = 2010
}

@book{eiben2015introduction,
  title={Introduction to evolutionary computing},
  author={Eiben, Agoston E and Smith, James E},
  year={2015},
  publisher={Springer}
}

@article{bartlett2023exhaustive,
  title={Exhaustive symbolic regression},
  author={Bartlett, Deaglan J and Desmond, Harry and Ferreira, Pedro G},
  journal={IEEE Transactions on Evolutionary Computation},
  year={2023},
  publisher={IEEE}
}

@article{willsey2021egg,
  title={Egg: Fast and extensible equality saturation},
  author={Willsey, Max and Nandi, Chandrakana and Wang, Yisu Remy and Flatt, Oliver and Tatlock, Zachary and Panchekha, Pavel},
  journal={Proceedings of the ACM on Programming Languages},
  volume={5},
  number={POPL},
  pages={1--29},
  year={2021},
  publisher={ACM New York, NY, USA}
}

@inproceedings{de2023reducing,
  title={Reducing Overparameterization of Symbolic Regression Models with Equality Saturation},
  author={de Franca, Fabricio Olivetti and Kronberger, Gabriel},
  booktitle={Proceedings of the Genetic and Evolutionary Computation Conference},
  pages={1064--1072},
  year={2023}
}

@article{kommenda2020parameter,
  title={Parameter identification for symbolic regression using nonlinear least squares},
  author={Kommenda, Michael and Burlacu, Bogdan and Kronberger, Gabriel and Affenzeller, Michael},
  journal={Genetic Programming and Evolvable Machines},
  volume={21},
  number={3},
  pages={471--501},
  year={2020},
  publisher={Springer}
}

@article{Guimera2020,
  title = {A {B}ayesian machine scientist to aid in the solution of challenging scientific problems},
  volume = {6},
  ISSN = {2375-2548},
  DOI = {10.1126/sciadv.aav6971},
  number = {5},
  journal = {Science Advances},
  publisher = {American Association for the Advancement of Science (AAAS)},
  author = {Guimerà,  Roger and Reichardt,  Ignasi and Aguilar-Mogas,  Antoni and Massucci,  Francesco A. and Miranda,  Manuel and Pallarès,  Jordi and Sales-Pardo,  Marta},
  year = {2020},
  month = jan 
}

@article{Reichardt2020,
  title = {Bayesian Machine Scientist to Compare Data Collapses for the {N}ikuradse Dataset},
  volume = {124},
  ISSN = {1079-7114},
  DOI = {10.1103/physrevlett.124.084503},
  number = {8},
  journal = {Physical Review Letters},
  publisher = {American Physical Society (APS)},
  author = {Reichardt,  Ignasi and Pallarès,  Jordi and Sales-Pardo,  Marta and Guimerà,  Roger},
  year = {2020},
  month = feb 
}

@techreport{nikuradse1950,
    author = {Nikuradse, Johann},
    title = {Laws of Flow in Rough Pipes},
    institution = {National Advisory Committee for Aeronautics Washington, NACA TM 1292 - Translation of "Strömungsgesetze in rauhen Rohren" VDI-Forschungsheft 361. Beilage zu “Forschung auf dem Gebiete des Ingenieurwesens" Ausgabe B Band 4, July/August 1933.},
    year = 1950
}

@inproceedings{daida2003makes,
  title={What makes a problem {GP}-hard? validating a hypothesis of structural causes},
  author={Daida, Jason M and Li, Hsiaolei and Tang, Ricky and Hilss, Adam M},
  booktitle={Genetic and Evolutionary Computation—GECCO 2003: Genetic and Evolutionary Computation Conference Chicago, IL, USA, July 12--16, 2003 Proceedings, Part II},
  pages={1665--1677},
  year={2003},
  organization={Springer}
}

@inproceedings{daida2003identifying,
  title={Identifying structural mechanisms in standard genetic programming},
  author={Daida, Jason M and Hilss, Adam M},
  booktitle={Genetic and Evolutionary Computation Conference},
  pages={1639--1651},
  year={2003},
  organization={Springer}
}

@inproceedings{gustafson2005improving,
  title={On improving genetic programming for symbolic regression},
  author={Gustafson, Steven and Burke, Edmund K and Krasnogor, Natalio},
  booktitle={2005 IEEE Congress on Evolutionary Computation},
  volume={1},
  pages={912--919},
  year={2005},
  organization={IEEE}
}

@inproceedings{ebner1999search,
  title={On the search space of genetic programming and its relation to {N}ature's search space},
  author={Ebner, Marc},
  booktitle={Proceedings of the 1999 Congress on Evolutionary Computation-CEC99 (Cat. No. 99TH8406)},
  volume={2},
  pages={1357--1361},
  year={1999},
  organization={IEEE}
}

@article{hu2018neutrality,
  title={Neutrality, robustness, and evolvability in genetic programming},
  author={Hu, Ting and Banzhaf, Wolfgang},
  journal={Genetic Programming Theory and Practice XIV},
  pages={101--117},
  year={2018},
  publisher={Springer}
}

@incollection{banzhaf2024combinatorics,
  title={How the Combinatorics of Neutral Spaces Leads Genetic Programming to Discover Simple Solutions},
  author={Banzhaf, Wolfgang and Hu, Ting and Ochoa, Gabriela},
  booktitle={Genetic Programming Theory and Practice XX},
  pages={65--86},
  year={2024},
  publisher={Springer}
}

@article{niehaus2007reducing,
  title={Reducing the number of fitness evaluations in graph genetic programming using a canonical graph indexed database},
  author={Niehaus, Jens and Igel, Christian and Banzhaf, Wolfgang},
  journal={Evolutionary Computation},
  volume={15},
  number={2},
  pages={199--221},
  year={2007},
  publisher={MIT Press One Rogers Street, Cambridge, MA 02142-1209, USA journals-info~…}
}

@inproceedings{hu2023phenotype,
  title={Phenotype search trajectory networks for linear genetic programming},
  author={Hu, Ting and Ochoa, Gabriela and Banzhaf, Wolfgang},
  booktitle={European Conference on Genetic Programming (Part of EvoStar)},
  pages={52--67},
  year={2023},
  organization={Springer}
}

@inproceedings{mcphee2008semantic,
  title={Semantic building blocks in genetic programming},
  author={McPhee, Nicholas Freitag and Ohs, Brian and Hutchison, Tyler},
  booktitle={Genetic Programming: 11th European Conference, EuroGP 2008, Naples, Italy, March 26-28, 2008. Proceedings 11},
  pages={134--145},
  year={2008},
  organization={Springer}
}

@inproceedings{burlacu2020hash,
  title={Hash-based tree similarity and simplification in genetic programming for symbolic regression},
  author={Burlacu, Bogdan and Kammerer, Lukas and Affenzeller, Michael and Kronberger, Gabriel},
  booktitle={Computer Aided Systems Theory--EUROCAST 2019: 17th International Conference, Las Palmas de Gran Canaria, Spain, February 17--22, 2019, Revised Selected Papers, Part I 17},
  pages={361--369},
  year={2020},
  organization={Springer}
}

@inproceedings{burlacu2019online,
  title={Online diversity control in symbolic regression via a fast hash-based tree similarity measure},
  author={Burlacu, Bogdan and Affenzeller, Michael and Kronberger, Gabriel and Kommenda, Michael},
  booktitle={2019 IEEE congress on evolutionary computation (CEC)},
  pages={2175--2182},
  year={2019},
  organization={IEEE}
}

@article{Vlcek2006,
title = {Shifted limited-memory variable metric methods for large-scale unconstrained optimization},
journal = {Journal of Computational and Applied Mathematics},
volume = {186},
number = {2},
pages = {365-390},
year = {2006},
issn = {0377-0427},
doi = {https://doi.org/10.1016/j.cam.2005.02.010},
author = {Jan Vlček and Ladislav Lukšan}
}

@inproceedings{burlacu2020operon,
  title={Operon {C}++ an efficient genetic programming framework for symbolic regression},
  author={Burlacu, Bogdan and Kronberger, Gabriel and Kommenda, Michael},
  booktitle={Proceedings of the 2020 Genetic and Evolutionary Computation Conference Companion},
  pages={1562--1570},
  year={2020}
}

@book{nelson1980techniques,
  title={Techniques for program verification},
  author={Nelson, Charles Gregory},
  year={1980},
  publisher={Stanford University}
}

@book{fletcher2000practical,
  title={Practical methods of optimization},
  author={Fletcher, Roger},
  year={2000},
  publisher={John Wiley \& Sons}
}

@inproceedings{worm2013prioritized,
  title={Prioritized grammar enumeration: symbolic regression by dynamic programming},
  author={Worm, Tony and Chiu, Kenneth},
  booktitle={Proceedings of the 15th annual conference on Genetic and evolutionary computation},
  pages={1021--1028},
  year={2013}
}

@article{kammerer2020symbolic,
  title={Symbolic regression by exhaustive search: Reducing the search space using syntactical constraints and efficient semantic structure deduplication},
  author={Kammerer, Lukas and Kronberger, Gabriel and Burlacu, Bogdan and Winkler, Stephan M and Kommenda, Michael and Affenzeller, Michael},
  journal={Genetic programming theory and practice XVII},
  pages={79--99},
  year={2020},
  publisher={Springer}
}

@article{rivero2022dome,
  title={Do{M}E: A deterministic technique for equation development and Symbolic Regression},
  author={Rivero, Daniel and Fernandez-Blanco, Enrique and Pazos, Alejandro},
  journal={Expert Systems with Applications},
  volume={198},
  pages={116712},
  year={2022},
  publisher={Elsevier}
}

@article{de2018greedy,
  title={A greedy search tree heuristic for symbolic regression},
  author={de Fran{\c{c}}a, Fabr{\'\i}cio Olivetti},
  journal={Information Sciences},
  volume={442},
  pages={18--32},
  year={2018},
  publisher={Elsevier}
}

@inproceedings{virgolin2017scalable,
  title={Scalable genetic programming by gene-pool optimal mixing and input-space entropy-based building-block learning},
  author={Virgolin, Marco and Alderliesten, Tanja and Witteveen, Cees and Bosman, Peter AN},
  booktitle={Proceedings of the Genetic and Evolutionary Computation Conference},
  pages={1041--1048},
  year={2017}
}

@inproceedings{de2022transformation,
  title={Transformation-interaction-rational representation for symbolic regression},
  author={de Fran{\c{c}}a, Fabr{\'\i}cio Olivetti},
  booktitle={Proceedings of the Genetic and Evolutionary Computation Conference},
  pages={920--928},
  year={2022}
}

@article{de2021interactionB,
  title={Interaction--transformation evolutionary algorithm for symbolic regression},
  author={de Franca, Fabricio Olivetti and Aldeia, Guilherme Seidyo Imai},
  journal={Evolutionary computation},
  volume={29},
  number={3},
  pages={367--390},
  year={2021},
  publisher={MIT Press One Rogers Street, Cambridge, MA 02142-1209, USA journals-info~…}
}

@article{kartelj2023rils,
  title={{RILS-ROLS}: robust symbolic regression via iterated local search and ordinary least squares},
  author={Kartelj, Aleksandar and Djukanovi{\'c}, Marko},
  journal={Journal of Big Data},
  volume={10},
  number={1},
  pages={71},
  year={2023},
  publisher={Springer}
}

@inproceedings{randall2022bingo,
  title={Bingo: a customizable framework for symbolic regression with genetic programming},
  author={Randall, David L and Townsend, Tyler S and Hochhalter, Jacob D and Bomarito, Geoffrey F},
  booktitle={Proceedings of the Genetic and Evolutionary Computation Conference Companion},
  pages={2282--2288},
  year={2022}
}

@inproceedings{cao2023genetic,
  title={Genetic Programming Symbolic Regression with Simplification-Pruning Operator for Solving Differential Equations},
  author={Cao, Lulu and Zheng, Zimo and Ding, Chenwen and Cai, Jinkai and Jiang, Min},
  booktitle={International Conference on Neural Information Processing},
  pages={287--298},
  year={2023},
  organization={Springer}
}

@article{seidyo2024inexact,
  title={Inexact Simplification of Symbolic Regression Expressions with Locality-sensitive Hashing},
  author={Seidyo Imai Aldeia, Guilherme and Olivetti de Franca, Fabricio and La Cava, William G},
  journal={arXiv e-prints},
  pages={arXiv--2404},
  year={2024}
}

@article{Langdon2021,
  title = {Genetic programming convergence},
  volume = {23},
  ISSN = {1573-7632},
  DOI = {10.1007/s10710-021-09405-9},
  number = {1},
  journal = {Genetic Programming and Evolvable Machines},
  publisher = {Springer Science and Business Media LLC},
  author = {Langdon,  W. B.},
  year = {2021},
  month = aug,
  pages = {71–104}
}

@article{kronberger_jsc_2024_to_appear,
  author = {Gabriel Kronberger and Fabricio Olivetti de Franca},
  title = {Effects of Reducing Redundant Parameters in Parameter Optimization for Symbolic Regression using Genetic Programming},
  journal = {Journal for Symbolic Computation},
  notes = {in review},
  year = {2024}
}

@ARTICLE{hansen2021coco,
    author = {Hansen, N. and Auger, A. and Ros, R. and Mersmann, O. and Tu{\v s}ar, T. and Brockhoff, D.},
    title = {{COCO}: A Platform for Comparing Continuous Optimizers in a Black-Box Setting},
    journal = {Optimization Methods and Software},
    doi = {10.1080/10556788.2020.1808977},
    pages = {114--144},
    issue = {1},
    volume = {36},
    year = 2021
    }

@misc{cranmerpysr,
  doi = {10.48550/ARXIV.2305.01582},
  author = {Cranmer,  Miles},
  title = {Interpretable Machine Learning for Science with PySR and SymbolicRegression.jl},
  publisher = {arXiv},
  year = {2023},
}
\end{document}